%% file: root.tex
\documentclass[letterpaper, 10 pt, conference]{ieeeconf}  
\let\labelindent\relax

\IEEEoverridecommandlockouts                              

\usepackage{graphicx} 
\usepackage{caption}
\usepackage{enumitem}
\usepackage{amsmath} 
\usepackage{amssymb}  
\usepackage{multirow}
\usepackage{booktabs}
\usepackage{color}
\usepackage{hyperref}

\title{\LARGE \bf
LanguageMPC: Large Language Models as Decision Makers for Autonomous Driving
}

\let\oldtwocolumn\twocolumn
\renewcommand\twocolumn[1][]{%
    \oldtwocolumn[{#1}{
    \begin{center}
           \includegraphics[scale=0.35]{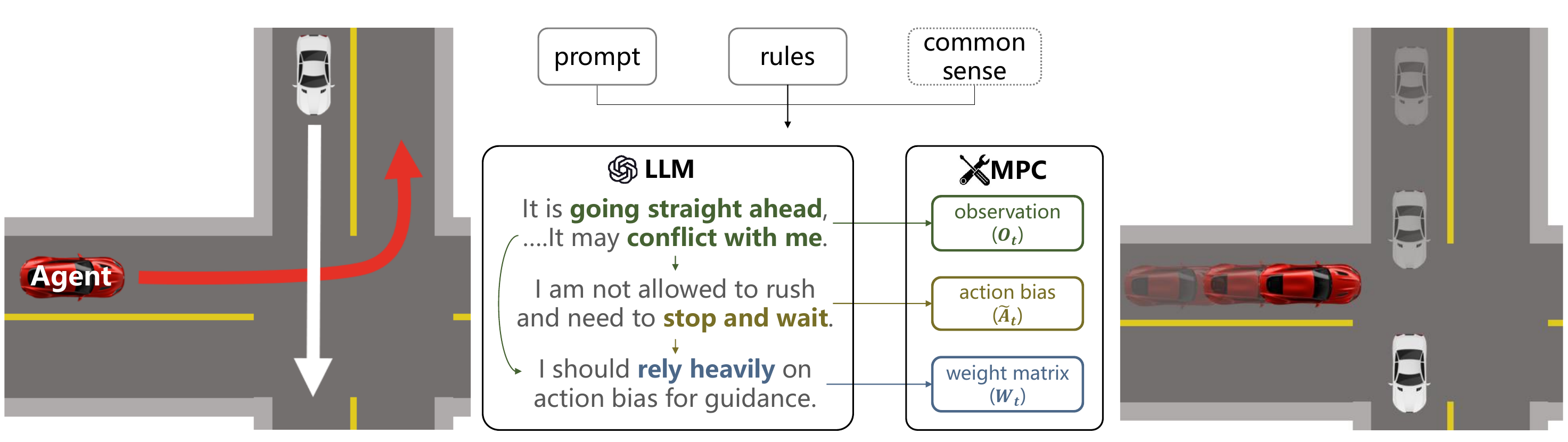}
           \captionof{figure}{For a left turn at an unsignalized intersection, the LLM makes a human-like decision to direct the MPC to slow down and wait, adhering to the traffic rule that left-turning vehicles must yield to oncoming traffic.}
           \label{fig:teaser}
        \end{center}
    }]
}

\author{
    Hao Sha$^{1}$*, Yao Mu$^{2}$*, Yuxuan Jiang$^{1}$, Guojian Zhan$^{1}$, Li Chen$^{2}$,
    Chenfeng Xu$^{3}$, Ping Luo$^{2}$, Shengbo Eben Li$^{1}$, \\ \textit{\small Senior Member, IEEE}, Masayoshi Tomizuka$^{3}$, \textit{\small Life Fellow, IEEE}, Wei Zhan$^{3}$, \textit{\small Member, IEEE}, Mingyu Ding$^{3}$, \textit{\small Member, IEEE}%
    \thanks{$^{1}$H. Sha, Y. Jiang, G. Zhan, and S. Li are with Tsinghua University.}%
    \thanks{$^{2}$Y. Mu, L. Chen, and P. Luo are with the University of Hong Kong.}%
    \thanks{$^{3}$C. Xu, M. Tomizuka, W. Zhan, and M. Ding are with the University of California, Berkeley.}%
    \thanks{*These authors contributed equally. Corresponding author: Mingyu Ding.}
}

\newcommand{\highlight}[1]{\textcolor{black}{#1}}

\begin{document}
\maketitle
\thispagestyle{empty}
\pagestyle{empty}

\begin{abstract}
Existing learning-based autonomous driving (AD) systems face challenges in comprehending high-level information, generalizing to rare events, and providing interpretability.
To address these problems, this work employs Large Language Models (LLMs) as a decision-making component for complex AD scenarios that require human commonsense understanding.
We devise cognitive pathways to enable comprehensive reasoning with LLMs, and develop algorithms for translating LLM decisions into actionable driving commands.
Through this approach, LLM decisions are seamlessly integrated with low-level controllers by guided parameter matrix adaptation.
Extensive experiments demonstrate that our proposed method not only consistently surpasses baseline approaches in single-vehicle tasks, but also helps handle complex driving behaviors even multi-vehicle coordination, thanks to the commonsense reasoning capabilities of LLMs.
This paper presents an initial step toward leveraging LLMs as effective decision-makers for intricate AD scenarios in terms of safety, efficiency, generalizability, and interoperability. We aspire for it to serve as inspiration for future research in this field.
\end{abstract}


\input{intro}
\input{related}
\input{method}
\input{exp}

\section{Conclusion}
This paper demonstrates the effectiveness of LLMs as a core component for high-level decision-making in AD systems. By integrating LLMs with MPC, our approach significantly improves performance in key areas such as safety, efficiency, and adaptability, particularly in complex and dynamic driving scenarios. The advanced reasoning capabilities and interpretability of LLMs help address limitations found in traditional learning-based methods, enhancing both transparency and flexibility. However, a limitation of our approach is that LLMs may not be able to respond quickly enough to sudden and highly time-critical events that occur on sub-second timescales. Addressing this challenge is a key direction for future work. We hope this work inspires continued exploration and innovation in applying LLMs to autonomous driving, paving the way for safer, more efficient, and adaptable transportation solutions.

\bibliography{IEEEabrv, IEEEexample}
\bibliographystyle{IEEEtran}




\end{document}

%% file: intro.tex
\vspace{-5pt}
\section{Introduction}
\vspace{-3pt}
Imagine you are behind the wheel, approaching an unsignalized intersection and planning to turn left, with an oncoming vehicle straight ahead. Human drivers instinctively know to slow down and yield according to traffic rules, even if it is possible to accelerate through the intersection. However, existing advanced learning-based Autonomous Driving (AD) systems typically require complex rules or reward function designs to handle such scenarios effectively~\cite{10138317,9351818}. This reliance on predefined rule bases often limits their ability to generalize to various situations.

Another challenge facing existing learning-based AD systems is the long-tail problem~\cite{9022290}. Both limited datasets and sampling efficiency~\cite{atakishiyev2023explainable} can present challenges for existing learning-based AD systems when making decisions in rare real-world driving scenarios. Chauffeurnet~\cite{bansal2018chauffeurnet} demonstrated such limits where even 30 million state-action samples were insufficient to learn an optimal policy that mapped bird's-eye view images (states) to control (action).
\begin{figure*}[t]
\centering
\includegraphics[scale=0.525]{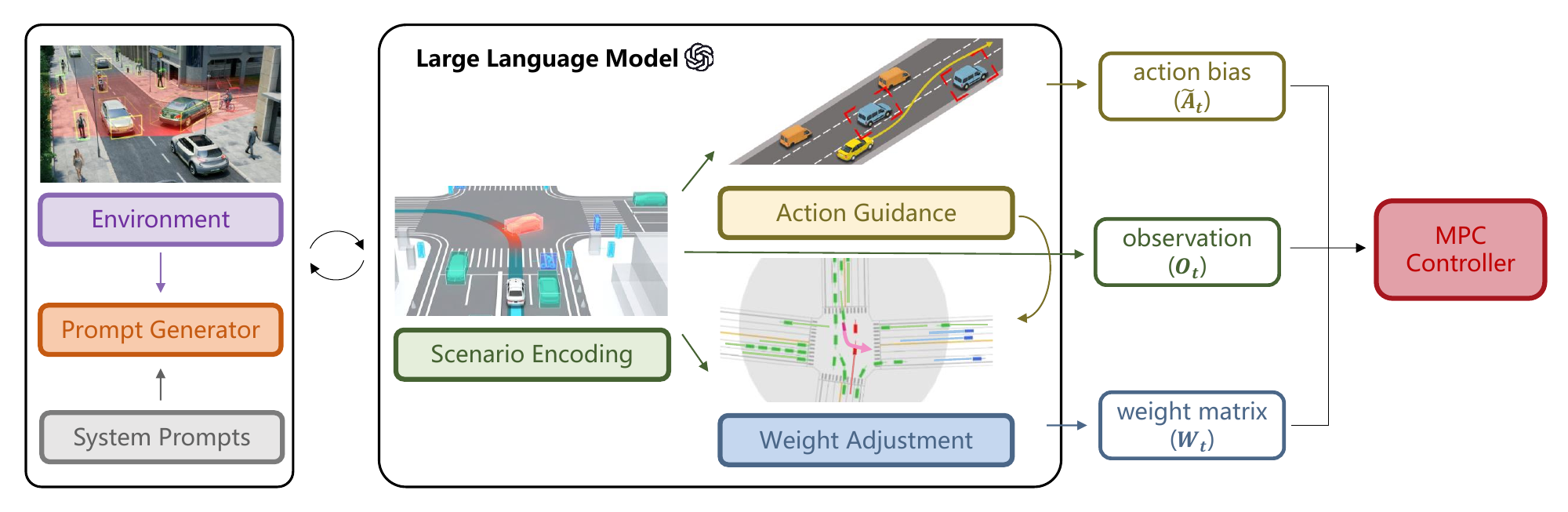}
\captionsetup{singlelinecheck=false}
\vspace{-16pt}
\caption{Pipeline of our system with LLM as the core of high-level decision-making. The LLM processes environment and system descriptions from the prompt generator to perform 1) scenario encoding, 2) action guidance, and 3) confidence adjustment, which dynamically adjust observation matrices $O_t$, action bias$\hat{A_t}$, and cost function weights $W_t$, guiding the MPC to execute appropriate control behaviors.}
\label{fig:pipeline}
\vspace{-10pt}
\end{figure*} 

\vspace{-1pt}
Furthermore, the lack of interpretability~\cite{gohel2021explainable} is a pressing issue for existing learning-based AD systems. A mature AD system must possess interpretability to gain recognition within society and regulatory entities, allowing it to be subject to targeted optimization and iterative improvements. Nevertheless, existing learning-based AD systems inherently resemble black boxes, making it challenging to discern their decision-making processes or understand the rationale behind their actions~\cite{atakishiyev2023explainable}. This lack of transparency can pose obstacles to the practical implementation of AD systems.

\vspace{-1pt}
Considering the aforementioned challenges, a fundamental question arises: \textit{Can we equip AD systems with the capability to think and drive like humans?} Our proposed solution involves employing a Large Language Model (LLM) to serve as the "brain" of the AD system.

Recent introductions of models like ChatGPT~\cite{chatgpt}, have positioned LLMs as having a certain level of human-like reasoning capabilities, owing to their innovative techniques such as Instruct Following and In-Context Learning (ICL)~\cite{dong2023survey}. LLMs can think like humans~\cite{fu2023drive}, and reason about new scenarios by combining common sense, and the visible thinking process makes them somewhat interpretable. These features make LLMs a powerful solution to the problems faced by AD systems described above.

In this paper, we leverage LLMs to analyze and reason about various scenarios, enabling it to provide high-level decisions, and by tuning parameter matrix, we convert high-level decisions into mathematical representations to guide the bottom-level controller, Model Predictive Control (MPC). Fig.~\ref{fig:teaser} illustrates the reasoning capabilities of our system for rare and complex scenarios, demonstrating its superiority in understanding high-level information, commonsense reasoning, and interpretability. Through quantitative experiments, we showcase that our system significantly surpasses existing learning-based and optimization-based methods for single-vehicle decision-making tasks, with Overall Cost decreasing by \textbf{18.1\%} and \textbf{16.4\%}. Additionally, through qualitative experiments, we demonstrate the capabilities of our system in addressing complex tasks, such as multi-vehicle joint control and driving behavior modulation guided by textual input.
The main contributions of this paper are as follows:
\vspace{-3pt}
\begin{enumerate}[leftmargin=0.5cm]
\itemsep 2pt
    \item We developed LanguageMPC, an end-to-end system that processes observations and outputs driving actions, guided by high-level, human-like decisions from the LLM. This integration enhances the understanding of complex scenarios and improves interpretability. 
    \item We develop a method that translates high-level textual decisions from the LLM into precise driving actions for the bottom-level controller, integrating the LLM's decision-making capabilities with the real-time control and robustness of the bottom-level controller.
    \item We demonstrate through quantitative experiments that LanguageMPC outperforms existing methods in key metrics. Additionally, it effectively manages complex tasks such as multi-vehicle coordination and driving behavior adjustment based on textual inputs. 
\end{enumerate}

%% file: related.tex
\section{Related Work}

\noindent \textbf{Autonomous Driving Autonomy.}
Modularity~\cite{8317797} plays an important role in many autonomous driving systems. It divides the complex autonomous driving task into multiple subproblems, including perception~\cite{li2023logonet}, planning~\cite{zhang2022rethinking}, and control~\cite{peng2018sim}, making each module relevant and somewhat interpretable. However, it also presents challenges in areas such as compatibility, error accumulation and inefficiency\cite{chib2023recent}. 
Recently, there was a large progress in end-to-end deep learning methods for autonomous systems~\cite{casas2021mp3,uniad,sadat2020perceive}. 
Though autonomous driving systems have achieved remarkable successes in planning and decision-making~\cite{zhang2022rethinking,10160326}, there are still problems in terms of interpretability~\cite{gohel2021explainable,chib2023recent}. At the same time, limitations in data and sampling efficiency~\cite{atakishiyev2023explainable} make it vulnerable to dealing with long-tail situations, especially interaction scenarios, in real-world environments~\cite{kong2023robo3d}.

\noindent \textbf{Large Language Models for Autonomous Driving.}
The remarkable achievements of LLMs are undeniably captivating, demonstrating LLM's human-like reasoning skills and generalization of human commonsense~\cite{bian2023chatgpt,ouyang2022training}.
In the autonomous driving industry, LLMs have the potential to revolutionize the paradigm of perception and decision-making. On the perceptual side, researchers combine visual encoders with LLM~\cite{radford2021learning} to enhance the AD system's understanding of visual high-level concepts contained in images~\cite{liu2023visual} and videos~\cite{zhang2023videollama}. Recent works such as DriveGPT4~\cite{xu2024drivegpt4}, HiLM-D~\cite{ding2023hilmd}, and Talk2BEV~\cite{choudhary2023talk2bev} have even demonstrated the powerful environment perception capabilities of LLMs enabled by vectorized visual embedding techniques.
In planning and decision-making, extensive research has shown that pre-trained LLMs integrate human common sense and help make human-like decisions~\cite{huang2023voxposer, cui2023drive, wen2024dilu}.
Also, the involvement of LLMs greatly enhances interpretability and safety, providing the potential to comprehend the transparency of the decision process. \cite{xu2024drivegpt4, cui2023drive, 10611018} showed that LLMs can reason about the transportation environment under the guidance of prompt and explain the decision process. 
In addition, LLMs show the potential to address interoperability challenges~\cite{chen2023driving, kamath2023new}. \cite{mao2023gptdriver, jin2023surrealdriver, keysan2023text} uses language as a bridge between multimodal data, enabling the union of perception and decision-making, demonstrating powerful understanding and reasoning.
However, the above work mainly focuses on demonstrating the perception, reasoning and high-level decision-making capabilities of LLMs, which are still lacking in generating final control actions.
While \cite{fu2023drive, mao2023gptdriver} can generate planning trajectories, \cite{fu2023drive} relies on predefined fixed rules, and \cite{mao2023gptdriver} still requires the output of LLMs to be decoded into trajectories using a deep learning model, and the decoder still faces the problem of being limited by the training set.
In this work, we combine the LLM directly with the controller MPC and use the powerful reasoning and decision-making capabilities of the LLM to guide the generation of the final control actions, allowing the system to drive like a human.


%% file: method.tex
\section{Methodology}

\begin{figure}[t]
\centering
\includegraphics[scale=0.435]{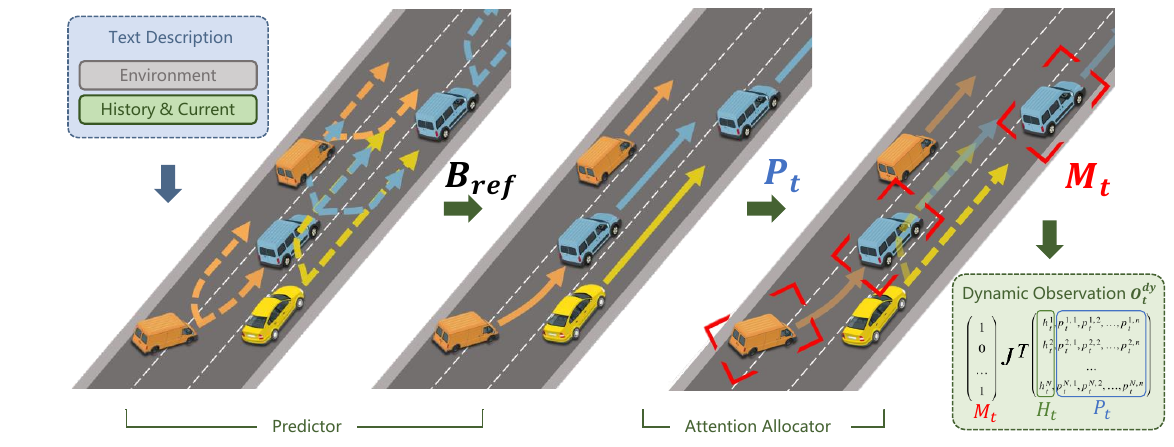}
\captionsetup{singlelinecheck=false}
\vspace{-16pt}
\caption{Scenario encoding. The yellow vehicle is the agent. Given three potential future trajectories for each vehicle, the LLM first selects the most likely trajectory to predict, representing it as $P_t$ using Bessel Curve control points. It then identifies the relevant traffic participants, setting those needing no attention to 0 via a binary mask $M_t$.}
\label{fig:encoder}
\vspace{-16pt}
\end{figure}  

We present an AD system that integrates the LLM with MPC to enhance decision-making in complex traffic scenarios. As illustrated in Fig.~\ref{fig:pipeline}, the LLM serves as the core component for high-level reasoning and decision-making, while the MPC acts as the low-level controller to execute decisions in real time.

The decision-making process begins with a prompt generator that translates real-time scenario data into descriptive text. This text includes details about static road elements, dynamic information about vehicles, and system prompts like traffic rules.
The LLM processes information through three sequential stages: 1) scenario encoding, 2) action guidance, and 3) weight adjustment. These stages collectively enable the system to accomplish scene understanding, reasoning, and high-level decision-making. Once these textual decisions are made, they are converted into mathematical representations: the observation matrix, action bias, and weight matrix. These elements guide bottom-level controller, the MPC, to give specific driving actions.

This framework integrates the advanced reasoning and decision-making capabilities of the LLM with the robustness and real-time control efficiency of the MPC:
\begin{enumerate}[leftmargin=0.5cm]
    \item The LLM functions as the "brain" of the AD system, enabling high-level reasoning and decision-making.
    \item The MPC implements the LLM's decisions under soft constraints, ensuring robust yet flexible control.
    \item Together, the LLM and MPC form a dual-frequency system that balances the longer inference time of the LLM with the real-time execution capabilities of the MPC, achieving effective real-time control.
\end{enumerate}

\vspace{-5pt}
\subsection{MPC Controller}
\label{sec:MPC}

The MPC solves a finite-time open-loop optimization problem online at each time step, based on the current measurement information obtained, and applies the initial control action from the optimal sequence to the vehicle.

The cost function of MPC is defined in the context of Markov Decision Process (MDP), which is commonly used to formulate vehicle control problems: $(S, A, C, P, p_0)$, where $S$ represents the state space, $A$ the action space, $C: S \times A \mapsto \mathbb{R}$ the cost function, $P: S \times A \mapsto S$ the state transition function, and $p_0$ the initial state distribution.  
The objective of the MPC is to find a sequence of actions $\mathbf{u}_{1:H} ={ \mathbf{u}_1, \dots, \mathbf{u}_H}$ that minimizes the expected cumulative cost $\small J(\mathbf{u}_{1:H}) = \sum_{t=1}^{H} C(\mathbf{s}_t, \mathbf{u}_t)$.

The cost function takes the following form:
\vspace{-5pt}
\begin{gather}
\small
C(\mathbf{s}, \mathbf{u}) = \sum_{i = 0}^M w_i \cdot \textrm{n}_i\big(r_i(\mathbf{s}, \mathbf{u}, \zeta_i)\big),
\end{gather}
where $w \in \mathbb{R}{+}$ are non-negative weights, $\textrm{n}(\cdot) : \mathbb{R} \rightarrow \mathbb{R}+$ is a twice-differentiable norm that reaches its minimum value at 0, and $r_i$ are residual terms parameterized by $\zeta_i$.
Designing a universal set of weights and residual terms for all driving scenarios is impractical due to the diversity of conditions encountered in real-world driving. Additionally, increasing the complexity of the cost function may reduce robustness~\cite{unknown}. Therefore, we use a simplified set of residual terms that incorporate action biases, adapting the weight matrices $\mathbf{W} = \left\{W_i\right\}$ based on LLM’s high-level decisions.

\begin{figure}[t]
\centering
\includegraphics[scale=0.32]{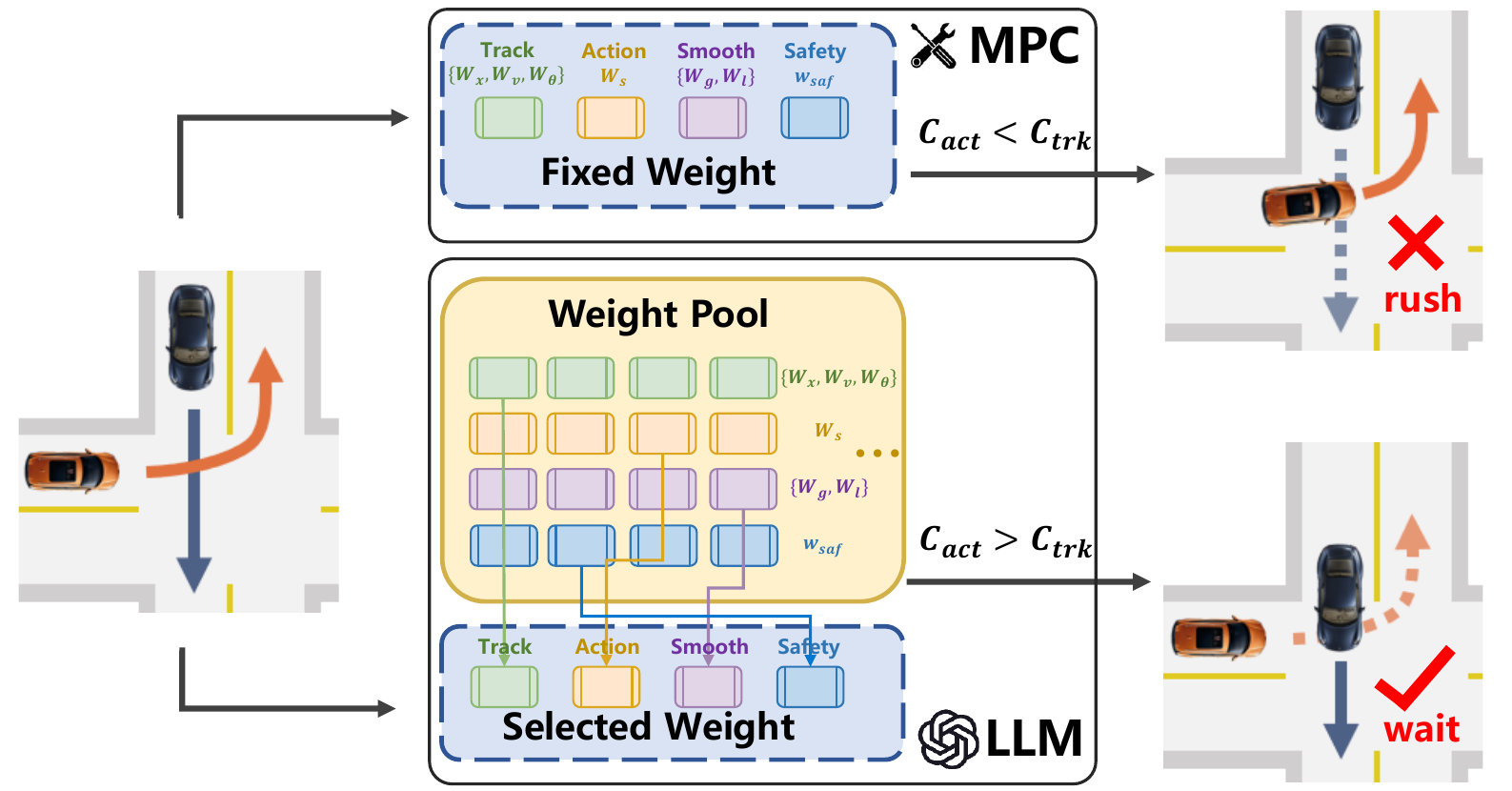}
\captionsetup{singlelinecheck=false}
\caption{Weight Adjustment. The LLM selects weights from the weight pool to balance the four components: tracking, action penalty, smoothness, and safety. In the example shown, the LLM selects a larger \( W_s \), making \( C_{\text{act}} > C_{\text{trk}} \), which prompts the MPC to prioritize following the LLM's decision to stop and wait over tracking the reference trajectory.}
\label{fig:weight}
\vspace{-12pt}
\end{figure}

The overall cost function is constructed as:
\begin{gather}
\small
C(\mathbf{s}, \mathbf{u}) = C_{trk}(\mathbf{s}, \mathbf{u}) + C_{act}(\mathbf{s}, \mathbf{u}) + C_{saf}(\mathbf{s}, \mathbf{u})
\end{gather}
We first generate a reference trajectory without considering signals and other traffic participants and then aim to follow that trajectory:
\begin{gather}
\small
C_{trk} = X^T W_{x} X + V^T W_v V + \Theta^T W_{\theta} \Theta,
\end{gather}
where $X = (x, y)^T$ represents the location, $V = (v_x, v_y)^T$ the velocity,
$\Theta = (\theta, \psi)^T$ with $\theta$ being the steering angle and $\psi$ the yaw angle.

To ensure smooth control, we impose limits on acceleration $a$ and steering angle $\theta$:
\vspace{-5pt}
\begin{gather}
\small
C_{act} = \overline{A}^T W_s \overline{A} + \nabla A^T W_g \nabla A + \nabla^2 A^T W_l \nabla^2 A
\vspace{-12pt}
\end{gather}
where $A = (a, \theta)^T$ and $\overline{A} = A - \hat{A}$, with $\hat{A}$ being the action bias given by LLM.

Lastly, we include a safety cost to maintain a safe distance from other vehicles:
\vspace{-8pt}
\begin{gather}
\small
C_{saf} = w_{saf} \cdot {\textstyle \sum_{i=0}^{N}} (d_i - \hat{d_i})^2,
\vspace{-5pt}
\end{gather}
where $d_i$ represents the actual distance to other vehicles, and $\hat{d}_i$ is the desired safety distance.

All weight matrices $W_i$ are diagonal. The LLM adapts the parameters $\mathbf{s}_t$, $\mathbf{W}$, and $\hat{A}$ for complex driving scenarios to guide the MPC controller.

\vspace{-5pt}
\subsection{Scenario Encoding}
\label{sec:attention}

After the processing of the prompt generator, the LLM obtains current and short-term historical information about the scenario.
This information is structured into an observation matrix, $\mathbf{O}_t = \{O^{st}_t, O^{dy}_t \}$, for the MPC to compute cost function, where $O^{st}_t$ represents static environmental elements and $O^{dy}_t$ represents dynamic information about surrounding vehicles.
In this section, we will describe how the LLM optimizes and modifies $\mathbf{O}_t$ to improve planning of MPC.

Effective decision-making requires not only understanding the current and historical states but also predicting future states. To generate these predictions, we first create potential reference trajectories based on $O_t^{st}$ for the ego vehicle using a Bessel curve:
\vspace{-10pt}
\begin{gather}
B_{ref} = \sum_{i=0}^n p_i 
\left(
\begin{matrix}
    i \\
    n
\end{matrix}
\right)
\gamma^i (1-\gamma)^{n-i}
\end{gather}
where $p_i$ represent the control points, including waypoint locations and the vehicle's position, and $\gamma \in \left[ 0, 1 \right]$. 
We use the points on the center lines of the three nearest lanes as control points and generate three Bessel curves. These three reference trajectories provide potential paths for the vehicle without considering other signals and traffic participants. Given these control points, the LLM then acts as a predictor to select the most suitable reference trajectory based on the scenario context.
This selected trajectory is used as the tracking target, as discussed in section \ref{sec:MPC}.
Next, the same process is applied to generate predicted trajectories for all traffic participants.
The dynamic observation matrix is then represented as a combination of historical data $H_t$ and these predicted trajectories $P_t\in \mathbb{R}^{N \times 2n}$, as $O^{dy}_t = (H_t, P_t) \in \mathbb{R}^{N \times C}$, where $N$ is the number of traffic participants.
In complex traffic scenarios, it is crucial to focus attention on relevant vehicles to avoid unnecessary computation and potential logical conflicts.
To achieve this, as shown on the right side of Fig. \ref{fig:encoder}, the LLM functions as an attention allocator. It identifies the vehicles that need to be prioritized and generates a binary mask $M_t \in \mathbb{R}^{N \times 1}$. This mask effectively filters out irrelevant vehicles by setting their corresponding data to zero in the observation matrix. The final dynamic observation matrix for the MPC is computed as:
\begin{gather}
O^{dy}_t = M_t J^T (H_t, P_t),
\end{gather}
where $J \in \mathbb{R}^{N \times 1}$ is a vector of ones. This approach ensures that the MPC focuses only on the relevant traffic participants, enabling efficient and effective decision-making. The MPC will then use the LLM-selected trajectory of the ego vehicle to calculate $C_{trk}$ for tracking and $C_{saf}$ using $O_t^{dy}$, optimizing the decision.

\begin{table}[t]
\centering
\captionsetup{singlelinecheck=false}
\caption{Evaluation of single-vehicle decision-making in various scenarios. SI: Signalized Intersection, USI: Unsignalized Intersection, RAB: Roundabouts, EOA: Emergency Obstacle Avoidance. Lower values indicate better performance.}
\vspace{-5pt}
\scalebox{.9}{
\begin{tabular}{ccccccccc}
\toprule   
\multirow{2}{*}{Scenario} & \multirow{2}{*}{Method} & \multirow{2}{*}{Col.} & \multirow{2}{*}{Fail} & \multirow{2}{*}{Ineff} & \multirow{2}{*}{Time} & \multicolumn{2}{c}{Penalty} & \multirow{2}{*}{O.C.}\\
\cmidrule(lr){7-8}
& & & & & & Acc & Dist\\
\midrule  
\multirow{6}{*}{SI} & MPC & 2 & 4 & 25.5 & 17.7 & 3.14 & 3.05 & \highlight{160.2}\\
& \highlight{DQN} & \highlight{6} & \highlight{0} & \highlight{34.0} & \highlight{14.3} & \highlight{3.90} & \highlight{2.98} & \highlight{173.6}\\
& \highlight{PPO} & \highlight{7} & \highlight{0} & \highlight{33.8} & \highlight{11.8} & \highlight{3.18} & \highlight{3.47} & \highlight{168.6}\\
& \highlight{SAC} & \highlight{6} & \highlight{0} & \highlight{30.1} & \highlight{\textbf{11.7}} & \highlight{3.66} & \highlight{3.51} & \highlight{172.8}\\
& ADP & 6 & 0 & 34.1 & 14.1 & 3.78 & 3.38 & \highlight{179.6}\\
& Ours & \textbf{0} & \textbf{0} & \textbf{13.9} & 25.7 & \textbf{1.31} & \textbf{1.20} & \highlight{\textbf{96.1}}\\
\midrule 
\multirow{6}{*}{USI} & MPC & 2 & 4 & 74.0 & 30.7 & 4.30 & 2.55 & \highlight{235.6}\\
& \highlight{DQN} & \highlight{7} & \highlight{1} & \highlight{58.3} & \highlight{28.6} & \highlight{6.06} & \highlight{3.53} & \highlight{262.7}\\
& \highlight{PPO} & \highlight{8} & \highlight{0} & \highlight{59.1} & \highlight{31.6} & \highlight{5.66} & \highlight{2.99} & \highlight{251.2}\\
& \highlight{SAC} & \highlight{7} & \highlight{1} & \highlight{80.8} & \highlight{30.8} & \highlight{4.71} & \highlight{3.21} & \highlight{261.9}\\
& ADP & 9 & \textbf{0} & 67.5 & \textbf{29.4} & 5.27 & 3.22 & \highlight{255.1}\\
& Ours & \textbf{0} & 1 & \textbf{33.7} & 42.2 & \textbf{1.94} & \textbf{0.98} & \highlight{\textbf{145.7}}\\
\midrule 
\multirow{6}{*}{Lane} & MPC & 0 & 0 & 4.1 & 6.8 & 0.20 & 0.08 & \highlight{18.9}\\
& \highlight{DQN} & \highlight{0} & \highlight{0} & \highlight{2.5} & \highlight{7.9} & \highlight{0.12} & \highlight{0.08} & \highlight{17.7}\\
& \highlight{PPO} & \highlight{0} & \highlight{0} & \highlight{2.0} & \highlight{7.6} & \highlight{0.15} & \highlight{0.10} & \highlight{17.7}\\
& \highlight{SAC} & \highlight{0} & \highlight{0} & \highlight{2.1} & \highlight{7.3} & \highlight{0.18} & \highlight{0.09} & \highlight{17.6}\\
& ADP & 0 & 0 & 2.3 & 6.8 & 0.15 & 0.09 & \highlight{16.5}\\
& Ours & \textbf{0} & \textbf{0} & \textbf{1.1} & \textbf{6.7} & \textbf{0.08} & \textbf{0.03} & \highlight{\textbf{13.0}}\\
\midrule 
\multirow{6}{*}{RAB} & MPC & 1 & 3 & 29.3 & 30.4 & 1.61 & 0.68 & \highlight{112.7}\\
& \highlight{DQN} & \highlight{6} & \highlight{0} & \highlight{31.6} & \highlight{31.2} & \highlight{1.58} & \highlight{0.67} & \highlight{115.5}\\
& \highlight{PPO} & \highlight{4} & \highlight{0} & \highlight{30.5} & \highlight{33.3} & \highlight{1.93} & \highlight{0.82} & \highlight{125.8}\\
& \highlight{SAC} & \highlight{5} & \highlight{0} & \highlight{30.2} & \highlight{32.5} & \highlight{1.81} & \highlight{0.79} & \highlight{121.9}\\
& ADP & 5 & 0 & 29.3 & \textbf{30.3} & 1.64 & 0.71 & \highlight{113.6}\\
& Ours & \textbf{0} & \textbf{0} & \textbf{26.8} & 31.9 & \textbf{1.51} & \textbf{0.65} & \highlight{\textbf{110.3}}\\
\midrule 
\multirow{6}{*}{EOA} & MPC & 8 & 4 & 33.4 & 17.3 & 3.34 & 2.06 & \highlight{150.7}\\
& \highlight{DQN} & \highlight{10} & \highlight{2} & \highlight{35.6} & \highlight{18.3} & \highlight{3.22} & \highlight{2.29} & \highlight{157.2}\\
& \highlight{PPO} & \highlight{10} & \highlight{1} & \highlight{30.3} & \highlight{16.8} & \highlight{3.13} & \highlight{2.13} & \highlight{145.1}\\
& \highlight{SAC} & \highlight{7} & \highlight{2} & \highlight{37.2} & \highlight{17.5} & \highlight{2.67} & \highlight{1.84} & \highlight{140.3}\\
& ADP & 11 & \textbf{0} & 32.3 & 16.9 & 2.99 & 1.96 & \highlight{141.7}\\
& Ours & \textbf{3} & 2 & \textbf{28.8} & \textbf{16.7} & \textbf{2.60} & \textbf{1.79} & \highlight{\textbf{128.7}}\\
\bottomrule    
\end{tabular}
}
\label{tab:t1}
\vspace{-12pt}
\end{table}

After the scenario encoding stage, LanguageMPC will further give action guidance. Since most LLMs have limited sensitivity to precise numerical values \cite{ahn2024large}, we discretize the action $A = (a, \theta)^T$ into a set of intervals. The midpoint of the LLM-selected interval is then used as action bias $\hat{A}$, which guides the MPC to generate control actions that align with the high-level decisions of the LLM.

\begin{figure*}[t]
\centering
\includegraphics[scale=0.65]{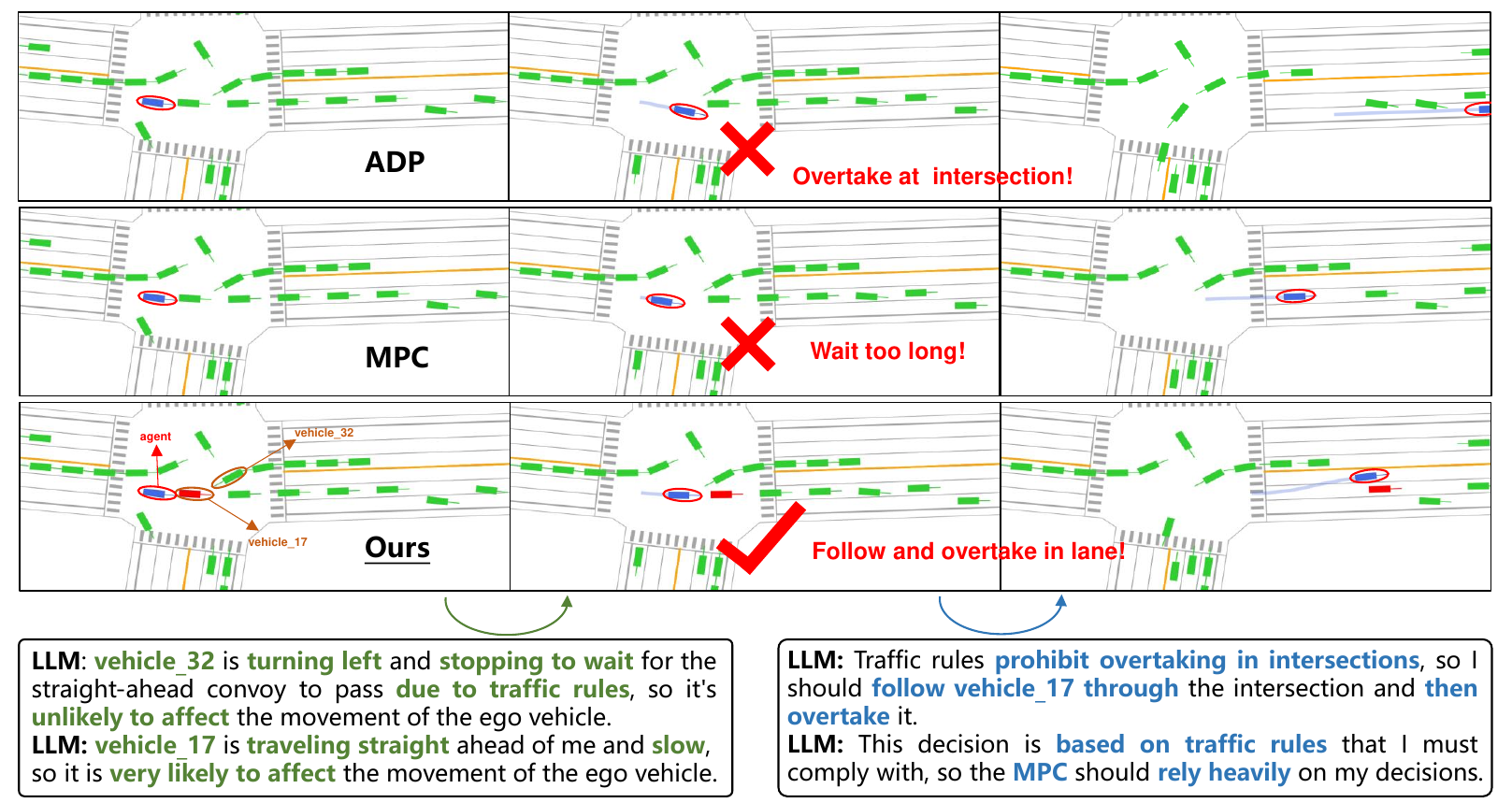}
\captionsetup{singlelinecheck=false}
\vspace{-17pt}
\caption{The ego vehicle is traveling straight through an unsignalized intersection. The red vehicle(s) in the last row represent those to which the LLM assigns attention, meaning information about these vehicles appears in the MPC dynamic observation matrix $O_t^{dy}$. This example demonstrates the LLM's ability to interpret high-level information, recognizing that "vehicle\_32" is yielding and focusing on the vehicle ahead to safely and compliantly navigate the intersection.}
\vspace{-13pt}
\label{fig:cross_straight}
\end{figure*} 

One challenge with MPC is its limited ability to generalize across diverse scenarios, as it relies on predefined weights $\mathbf{W} = {W_i}$~\cite{unknown}.
In our approach, as detailed in Section~\ref{sec:MPC}, the weight matrices $\mathbf{W} = \left\{W_x, W_v, W_{\theta}, W_s, W_g, W_l, w_{safe}\right\}$ are categorized into four components: tracking $\left\{W_x, W_v, W_{\theta} \right\}$, action panelty $W_s$, smoothness $\left\{W_g, W_l \right\}$, and safety $w_{saf}$. Unlike fixed weight matrices designed for all scenarios, LanguageMPC dynamically selects weights from a predefined pool for each specific scenario, as shown in Fig.~\ref{fig:weight}. These weights directly influence the cost function, guiding the MPC to adjust the trade-offs among the four components. This approach allows the MPC to adapt to different situations while ensuring robust control without relying on direct action commands from the LLM.

\vspace{-8pt}
\subsection{Dual-frequency System}
\vspace{-2pt}
\label{sec:real-time}

The decision-making speed of the LLM in LanguageMPC is relatively slow due to its large number of parameters. To address this, we implement a dual-frequency system inspired by DriveVLM \cite{tian2024drivevlm}. In this system, we separate decision-making into two levels: low-frequency for high-level decisions and high-frequency for trajectory planning and control. At the low-frequency level, the LLM makes strategic decisions every few seconds. These strategic decisions are sufficient to adapt to changes in driving conditions that occur over a longer timescale. In contrast, the MPC operates at the high-frequency level, performing fine-grained trajectory planning and control between LLM decisions. It generates real-time control actions based on the high-level guidance provided by the LLM. This dual-frequency approach balances the slower, computationally intensive inference of the LLM with the real-time demands of vehicle control, allowing LanguageMPC to achieve robust performance in dynamic driving scenarios.

%% file: exp.tex
\vspace{-5pt}
\section{Experiments}

In this section, we evaluate the performance of LanguageMPC, using GPT-3.5 as the base LLM model and LangChain as the manager of LLM text output. We assess its reasoning and decision-making capabilities through a series of experiments designed to test its effectiveness in single-vehicle decision-making and its adaptability to complex scenarios, such as driving behavior modulation and multi-vehicle coordination. We conduct quantitative experiments to measure its performance on various driving tasks and qualitative analyses to showcase its ability to handle challenging situations, including rare corner cases and coordinated actions between multiple vehicles.
\vspace{-5pt}
\subsection{Settings}

\noindent \textbf{Simulator}. We conduct simulation experiments using the SUMO simulator \cite{sumo} with IDSim \cite{10324473} to generate dynamic traffic scenarios, including vehicles, bicycles, and pedestrians. Five distinct driving scenarios are evaluated: Signalized Intersections, Unsignalized Intersections, Lanes, Roundabouts, and Emergency Obstacle Avoidance. Each scenario is initialized with varying road structures, traffic densities, and participant behaviors to simulate challenging conditions. 
Traffic parameters (densities, positions, speeds, target positions, aggressiveness) are randomly initialized and simulated with the IDSim agent. From these generated scenarios, we exclude straightforward situations, such as driving straight within a lane, and instead focus on complex corner cases, with 25 different scenes reserved for each scenario. For Emergency Obstacle Avoidance, low-speed vehicles are randomly placed in front of the ego vehicle to simulate sudden obstacles.

\noindent \textbf{Baselines}. We compare LanguageMPC against two baselines: \textbf{1) Model Predictive Control (MPC)} \cite{9760270}: We use a traditional MPC system with a fixed weight matrix, consistent with the control module in LanguageMPC. The dynamic observation matrix $O_t^{dy}$ incorporates all traffic participants within the sensing range. To predict the movement of these participants, all waypoints are used as control points for generating reference trajectories. The trajectory closest to each participant’s current path is chosen as the predicted trajectory. This baseline has been finely tuned and validated using real-world vehicle experiments conducted at iDLab, Tsinghua University. 
\textbf{2) Reinforcement Learning-Based Planning (RL)}: We employ four RL methods: Deep Q-Network (DQN) \cite{mnih2013playingatarideepreinforcement}, Proximal Policy Optimization (PPO) \cite{schulman2017proximalpolicyoptimizationalgorithms}, Soft Actor-Critic (SAC) \cite{haarnoja2018softactorcriticoffpolicymaximum}, and Finite-Horizon Approximate Dynamic Programming (ADP)\cite{9760270,9857655}.
For all RL methods, the MPC observation matrix $O_t$ is used as input to produce the control action $A = (a, \theta)^T$.
Training is performed on IDSim across the five types of randomly generated scenarios, with a total of $2 \times 10^5$ epochs.


\vspace{-5pt}
\subsection{Metrics}
\vspace{-3pt}

\begin{table}[t]
\centering
\captionsetup{singlelinecheck=false}
\caption{Ablation experiments. $O^{dy}$ indicates the use of the scenario encoder to modify the MPC observation matrix, while $\hat{A} \& W$ indicates the use of the LLM for action bias $\hat{A}$ and weight matrix $\mathbf{W}$ generation.}
\vspace{-5pt}
\scalebox{.81}{
\begin{tabular}{cccccccccc}
\toprule   
\multirow{2}{*}{Scenario} & \multirow{2}{*}{$O^{dy}$} & \multirow{2}{*}{$\hat{A} \& W$} & \multirow{2}{*}{Coll.} & \multirow{2}{*}{Fail} & \multirow{2}{*}{Ineff} & \multirow{2}{*}{Time} & \multicolumn{2}{c}{Penalty} & \multirow{2}{*}{O.C.}\\
\cmidrule(lr){8-9}
& & & & & & & Acc & Dist & \\
\midrule  
\multirow{4}{*}{USI} & $\times$ & $\times$ & 2 & 4 & 74.0 & 30.7 & 4.30 & 2.55 & \highlight{235.6}\\
& \checkmark & $\times$ & 2 & 3 & 69.8 & \textbf{28.7} & 3.95 & 2.32 & \highlight{218.5}\\
& $\times$ & \checkmark & \textbf{0} & 2 & 42.0 & 44.0 & 2.37 & 1.21 & \highlight{167.8}\\
& \checkmark & \checkmark & \textbf{0} & \textbf{1} & \textbf{33.7} & 42.2 & \textbf{1.94} & \textbf{0.98} & \highlight{\textbf{145.7}}\\
\midrule 
\multirow{4}{*}{RAB} & $\times$ & $\times$ & 1 & 3 & 29.3 & 30.4 & 1.61 & 0.68 & \highlight{112.7}\\
& \checkmark & $\times$ & 1 & 3 & 29.4 & \textbf{30.3} & 1.61 & 0.69 & \highlight{112.8}\\
& $\times$ & \checkmark & 0 & 0 & 27.8 & 30.9 & 1.53 & 0.67 & \highlight{110.5}\\
& \checkmark & \checkmark & \textbf{0} & \textbf{0} & \textbf{26.8} & 31.9 & \textbf{1.51} & \textbf{0.65} & \highlight{\textbf{110.3}}\\
\bottomrule    
\end{tabular}
}
\vspace{-12pt}
\label{tab:t2}
\end{table}

\noindent \textbf{Failure and Collision Cases (Fail/Col.):} We track failure cases where the ego vehicle fails to reach its target area within a 200-second time frame. Collision cases are recorded separately to monitor safety incidents.

\noindent \textbf{Inefficiency (Ineff):} We measure traffic flow efficiency by calculating the average speed difference between the ego vehicle’s maximum speed and its current speed:
\vspace{-5pt}
\begin{gather}
P_{eff} = \frac{1}{N} {\textstyle \sum_{i = 1}^N} (v_i^{max}-v_i)
\end{gather}
This metric focuses on the lead vehicle within each convoy, as it is most affected by the ego vehicle's driving behavior. Vehicles stopped at red lights are excluded.

\noindent \textbf{Time Efficiency (Time):} This metric records the time taken by the ego vehicle to reach its target area, reflecting the overall driving efficiency.

\noindent \textbf{Safety Penalty (Penalty):} The penalty metric assesses the safety of the ego vehicle’s driving behavior.It considers both the distance between the ego vehicle and surrounding vehicles, and the deceleration of those vehicles. Smaller distances and higher deceleration rates indicate more unsafe driving and contribute to a higher penalty score:
\vspace{-5pt}
\begin{equation}
P_{acc} = \sum_{i = 1}^{N}  ReLU (d_i - d_0), P_{dist} = \sum_{i = 1}^N d_i f(a_0-a_i)
\end{equation}
where $f$ is a step function from 0 to 1 with $a_0-a_i=0$ as the cutoff.

\noindent
\textbf{Overall Cost (O.C.):} To provide a comprehensive evaluation, we combine the above metrics into an overall cost function using weighted values:
\vspace{-5pt}
\begin{gather}
{\rm O.C.} = P_{eff} + 1.5t + 15 P_{acc} + 20 P_{dist}
\vspace{-5pt}
\end{gather}
These weights balance the influence of each metric, ensuring that each contributes equally by matching their magnitudes.

\vspace{-5pt}
\subsection{Single-vehicle Decision-making}
\label{sec:single}

The quantitative results in Table~\ref{tab:t1} show significant reductions in overall cost across all scenarios, including Emergency Obstacle Avoidance, highlighting our system's superior driving performance. It reports minimal failures and no collisions, demonstrating safety and reliability.

In intersection and roundabout scenarios, our method slightly increases travel time but improves traffic flow efficiency and reduces safety penalties, reflecting cautious, regulation-compliant driving. In lane-changing and Emergency Obstacle Avoidance scenarios, it outperforms RL and MPC across all metrics, excelling in complex and dynamic situations.

\noindent \textbf{Ablation Study.} To gain deeper insights into the system’s capabilities, we performed ablation experiments in two challenging driving scenarios: unsignalized intersections and roundabouts, as shown in Table~\ref{tab:t2}. These ablations highlight the contributions of each component of LanguageMPC. When using only the scenario encoder, significant improvements are observed in intention prediction and attention allocation compared to baseline MPC. In contrast, using only action bias and weight matrix adjustments leads to substantial improvements in all metrics except time. The full configuration, integrating both components, achieves the best performance, showcasing the synergistic effect of scenario understanding and high-level decision-making.

\begin{figure}[t]
\centering
\includegraphics[scale=0.385]{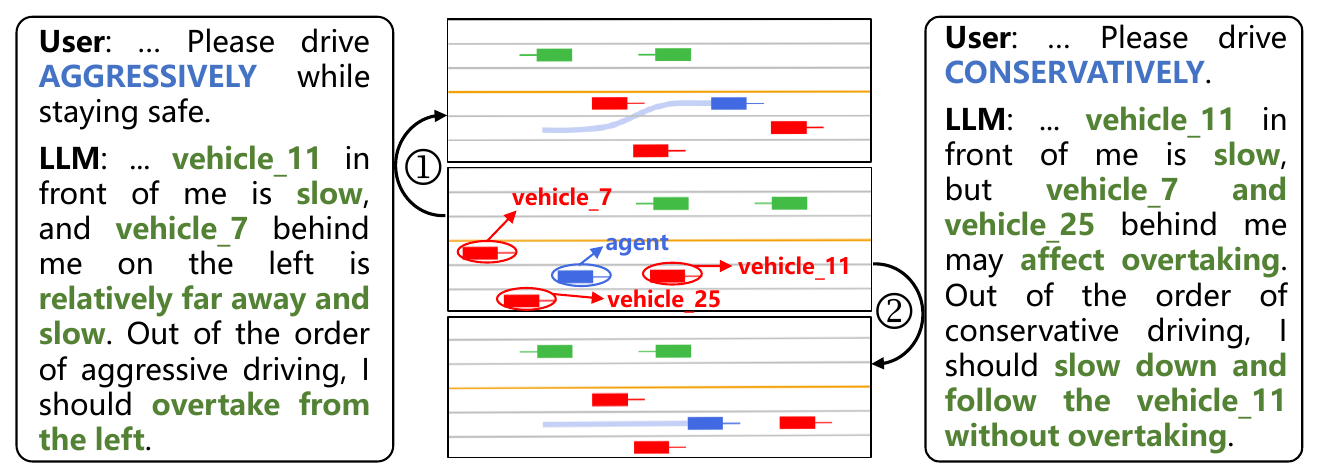}
\captionsetup{singlelinecheck=false}
\vspace{-16pt}
\caption{The ego vehicle adjusts its driving style based on user instructions. When directed to drive conservatively, it slows down and follows the vehicle ahead. Under aggressive driving commands, it safely overtakes the slower vehicle.}
\label{fig:style}
\vspace{-14pt}
\end{figure}

Fig.~\ref{fig:cross_straight} illustrates how LanguageMPC components work together. The ego vehicle navigates an unsignalized intersection, showcasing the system's ability to predict surrounding vehicles' intentions, allocate attention effectively, and make decisions aligned with traffic regulations.

\noindent \textbf{1) Intention Prediction and Attention Allocation.} The left side of Fig.~\ref{fig:cross_straight} highlights LanguageMPC's scenario encoding. Unlike the MPC, which fails to anticipate "vehicle\_32" yielding, LanguageMPC correctly interprets its intent and focuses on the vehicle ahead that affects the ego vehicle’s path. This demonstrates the LLM’s ability to understand traffic dynamics and allocate attention for rational driving behavior.

\noindent \textbf{2) High-level Reasoning.} In the same scenario, the RL method incorrectly overtakes within the intersection, violating traffic rules. LanguageMPC, however, understands traffic regulations and chooses to follow the vehicle until exiting the intersection before overtaking, as shown in the dialogue box. This demonstrates the LLM’s ability to reason about high-level constraints and make rule-compliant decisions in complex scenarios.

\vspace{-5pt}
\subsection{Text-modulated Driving}
\vspace{-2pt}

\noindent \textbf{Driving Style Adjustment.} Users often wish to customize AD systems to match their preferences for efficiency and comfort. Traditional systems require complex rule or reward function designs~\cite{chang2023editing}, while our approach simplifies this by enabling users to provide simple textual descriptions. As shown in Fig.~\ref{fig:style}, when instructed to drive conservatively, LanguageMPC slows down and follows the car ahead, whereas an aggressive style leads to safe overtaking.

\begin{figure}[t]
\centering
\includegraphics[scale=0.5]{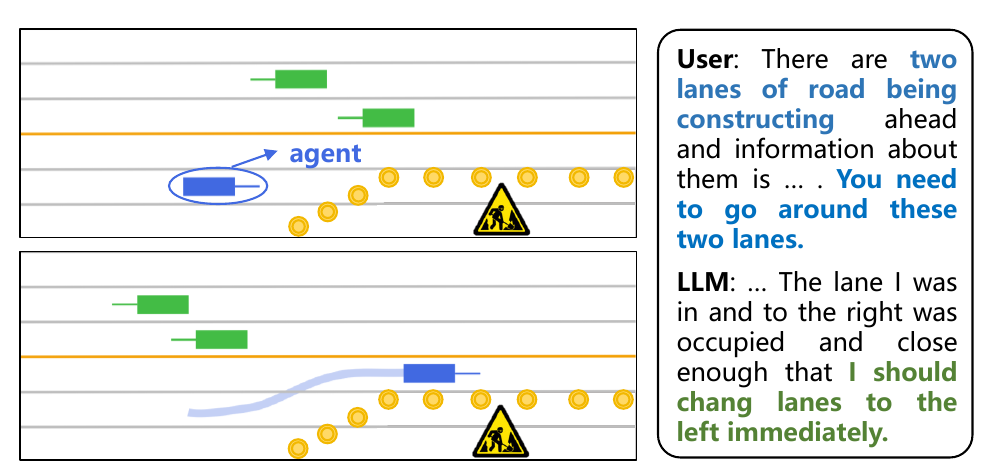}
\captionsetup{singlelinecheck=false}
\vspace{-6pt}
\caption{The LanguageMPC changes lanes under textual guidance to safely navigate a road construction zone, showcasing the system’s ability to follow user instructions and handle complex driving scenarios effectively.}
\label{fig:txt}
\vspace{-14pt}
\end{figure}

\noindent \textbf{Textual Guidance for Complex Scenarios.} Complex driving situations, such as road construction, pose challenges for many AD systems~\cite{10138317}. Our approach addresses this by enabling users to provide textual instructions to guide decision-making. As shown in Fig.~\ref{fig:txt}, the LLM used textual guidance to change lanes and avoid the blocked area, allowing the system to adapt effectively to unusual or dynamic scenarios.

\vspace{-5pt}
\subsection{Multi-vehicle Joint Control} 
\vspace{-2pt}

Multi-vehicle joint control enhances transportation efficiency and safety in dynamic environments. Traditional centralized and distributed methods struggle in unpredictable scenarios~\cite{10172912}. To address this, we propose a hybrid solution: each vehicle is controlled by a distributed LLM, while a central LLM coordinates the convoy. Each distributed LLM shares its local status with the central LLM, which then issues coordinated commands for adaptive decision-making.

\begin{figure}[t]
\centering
\includegraphics[scale=0.305]{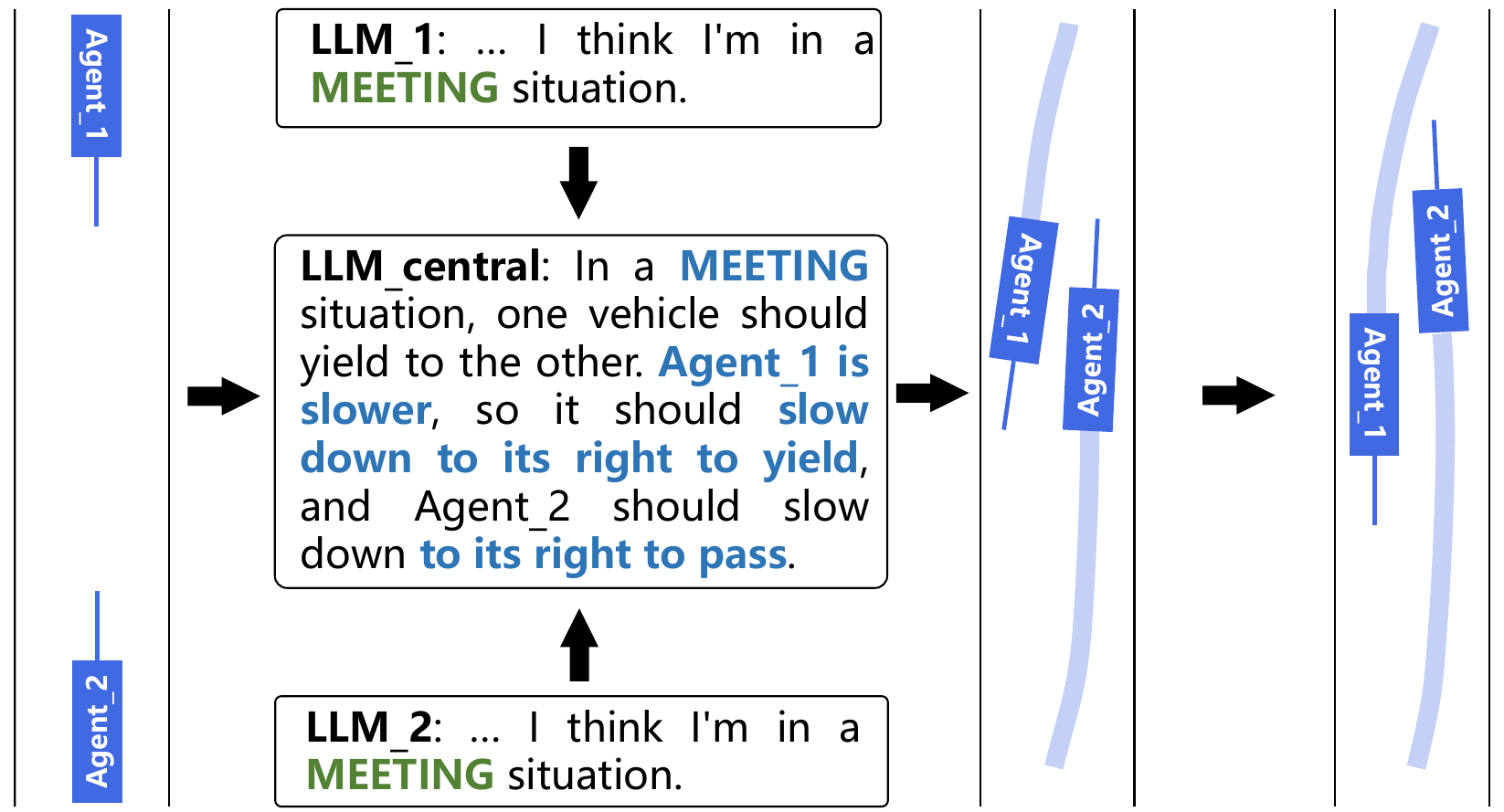}
\captionsetup{singlelinecheck=false}
\vspace{-6pt}
\caption{Two vehicles meet on a narrow road. Distributed LLMs detect the situation and inform the central LLM, which coordinates their actions, with one yielding and the other proceeding, demonstrating effective multi-vehicle coordination.}
\label{fig:multi}
\vspace{-14pt}
\end{figure}

Fig.~\ref{fig:multi} illustrates a corner case: a narrow road encounter between two vehicles. For a single-vehicle AD system, this scenario requires complex decision-making. However, our approach simplifies it by leveraging both centralized and distributed LLMs. In this case, the distributed LLMs recognize the situation, report their statuses to the central LLM, which then coordinates actions—one vehicle decelerates while the other proceeds. This showcases the effective collaboration between centralized decision-making and distributed control for optimized multi-vehicle coordination.

\vspace{-5pt}
\subsection{LLM Inference Delay}
\vspace{-2pt}

We evaluated the inference delay of LLMs and its impact on decision-making performance. In our test scenario, the number of tokens inferred by the LLM for a single decision ranged from 2k to 4k, resulting in an average inference delay of approximately 2.5 seconds. Notably, we implemented a rule-based pre-evaluation process to assess emergency situations, where the LLM bypasses the scenario encoding step and directly provides action guidance and adjusts weights. This modification significantly reduced the inference time during emergencies. Specifically, the number of tokens for Emergency Obstacle Avoidance dropped to under 1.5k, and the reasoning delay was reduced to less than 1.5 seconds.
In practice, high-level driving decisions typically evolve slowly and coherently, meaning the inference speed of the LLM is sufficient for most scenarios. Furthermore, LLM constraints on MPC are soft constraints that affect the cost function, rather than directly modifying actions. This ensures that the MPC remains robust despite the low-frequency decisions made by the LLM.